\pdfoutput=1

\documentclass[11pt]{article}

\usepackage[]{naacl2021}

\usepackage{times}
\usepackage{latexsym}
\usepackage{graphicx}
\usepackage{svg}
\usepackage{multirow}
\usepackage{amssymb}
\usepackage{pifont}

\newcommand{\cmark}{\ding{51}}
\newcommand{\xmark}{\ding{55}}

\usepackage[T1]{fontenc}

\usepackage[utf8]{inputenc}

\usepackage{microtype}

\title{Neural reality of argument structure constructions}

\author{Bai Li$^{1,4}$, Zining Zhu$^{1,4}$ Guillaume Thomas$^{2}$, Frank Rudzicz$^{1,4,5}$, Yang Xu$^{1,3,4}$ \\
$^1$ University of Toronto, Department of Computer Science \\
$^2$ University of Toronto, Department of Linguistics \\
$^3$ University of Toronto, Cognitive Science Program \\
$^4$ Vector Institute for Artificial Intelligence \hspace{0.5em} $^5$ Unity Health Toronto \\
\texttt{\{bai, zining, frank, yangxu\}@cs.toronto.edu} \\ 
\texttt{guillaume.thomas@utoronto.ca}
}

\begin{document}
\maketitle
\begin{abstract}
In lexicalist linguistic theories, argument structure is assumed to be predictable from the meaning of verbs. As a result, the verb is the primary determinant of the meaning of a clause. In contrast, construction grammarians propose that argument structure is encoded in constructions (or form-meaning pairs) that are distinct from verbs. Decades of psycholinguistic research have produced substantial empirical evidence in favor of the construction view. Here we adapt several psycholinguistic studies to probe for the existence of argument structure constructions (ASCs) in Transformer-based language models (LMs). First, using a sentence sorting experiment, we find that sentences sharing the same construction are closer in embedding space than sentences sharing the same verb. Furthermore, LMs increasingly prefer grouping by construction with more input data, mirroring the behaviour of non-native language learners. Second, in a ``Jabberwocky'' priming-based experiment, we find that LMs associate ASCs with meaning, even in semantically nonsensical sentences. Our work  offers the first evidence for ASCs in LMs and highlights the potential to devise novel probing methods grounded in psycholinguistic research.
\end{abstract}

\section{Introduction}

Pretrained Transformer-based language models (LMs) such as BERT \citep{bert} and RoBERTa \citep{roberta} have recently achieved impressive results on many natural language tasks, spawning a new interdisciplinary field of aligning LMs with linguistic theory and probing the linguistic capabilities of LMs \citep{linzen-baroni}. Most probing work so far has investigated the linguistic knowledge of LMs on phenomena such as agreement, binding, licensing, and movement \citep{blimp, hu-syntax-assessment} with a particular focus on determining whether a sentence is linguistically acceptable \citep{schutze1996}. Relatively little work has attempted to determine whether the linguistic knowledge induced by LMs is more similar to a formal grammar of the sort postulated by mainstream generative linguistics \citep{chomsky1965, chomsky1981, chomsky1995}, or to a network of form-meaning pairs as advocated by construction grammar \citep{goldberg1995, goldberg2006}.

\begin{figure}
    \centering
    \includegraphics[width=1.0\linewidth]{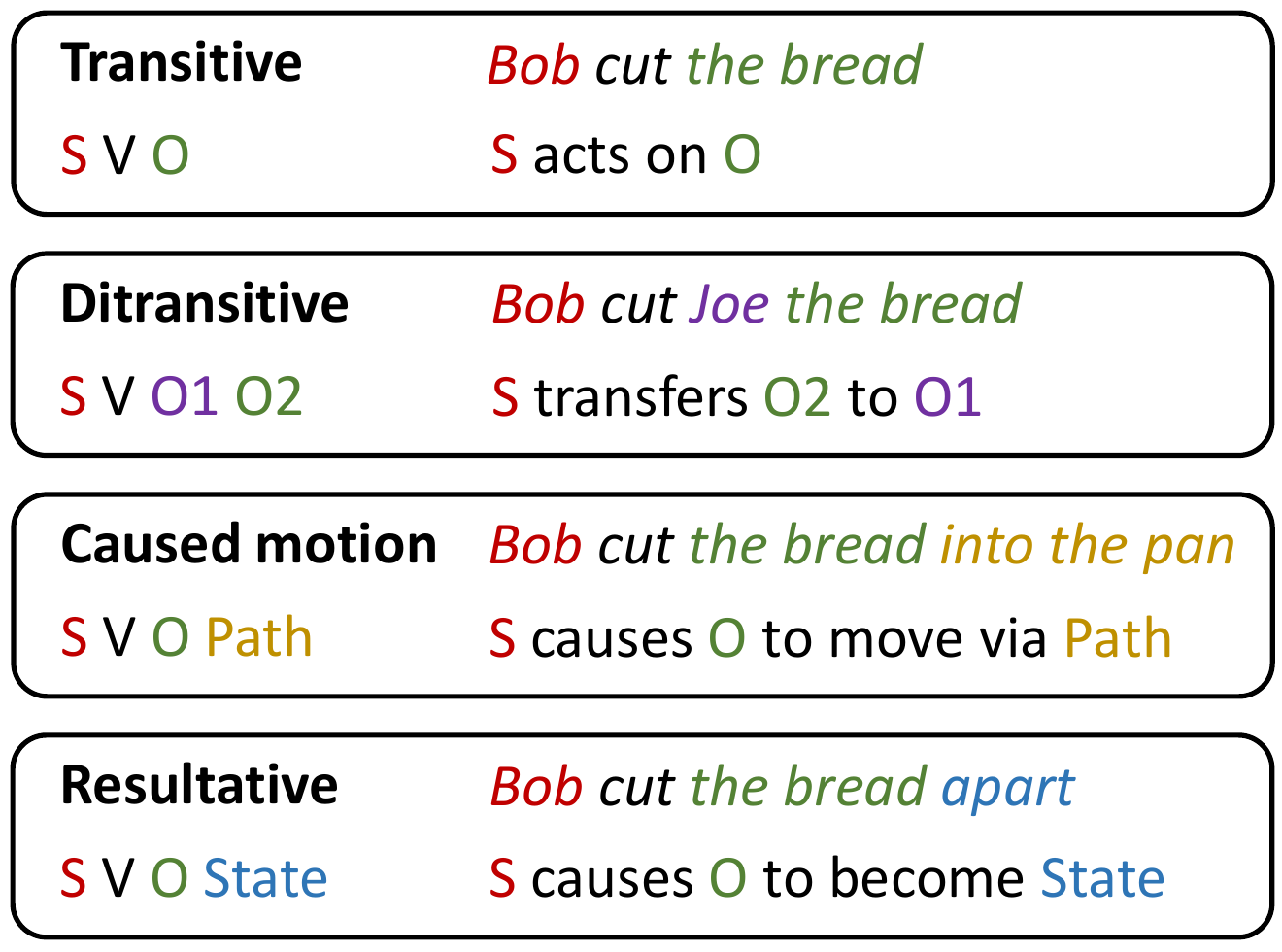}
    \caption{Four argument structure constructions (ASCs) used by \citet{bencini-goldberg}, with example sentences (top right). Constructions are mappings between form (bottom left) and meaning (bottom right).}
    \label{fig:cxn-examples}
\end{figure}

One area where construction grammar disagrees with many generative theories of language is in the analysis of the argument structure of verbs, that is, the specification of the number of arguments that a verb takes, their semantic relation to the verb, and their syntactic form \citep{levin-rappaport2005}. Lexicalist theories were long dominant in generative grammar \citep{chomsky1981, kaplan-bresnan1982, pollard-sag1987}. In lexicalist theories, argument structure is assumed to be encoded in the lexical entry of the verb: for example, the verb {\em visit} is lexically specified as being transitive and as requiring a noun phrase object \citep{chomsky1986}. In contrast, construction grammar suggests that argument structure is encoded in form-meaning pairs known as {\em argument structure constructions} (ASCs, Figure \ref{fig:cxn-examples}), which are distinct from verbs. The argument structure of a verb is determined by pairing it with an ASC \citep{goldberg1995}. To date, a substantial body of psycholinguistic work has provided evidence for the psychological reality of ASCs in sentence sorting \citep{bencini-goldberg, gries-wulff}, priming \citep{ziegler19}, and novel verb experiments \citep{kaschak-glenberg, johnson-goldberg}.

Here we connect basic research in ASCs with neural probing by adapting several psycholinguistic studies to Transformer-based LMs and show evidence for the neural reality of ASCs. Our first case study is based on sentence sorting \citep{bencini-goldberg}; we discover that in English, German, Italian, and Spanish, LMs consider sentences that share the same construction to be more semantically similar than sentences sharing the main verb. Furthermore, this preference for constructional meaning only manifests in larger LMs (trained with more data), whereas smaller LMs rely on the main verb, an easily accessible surface feature. Human experiments with non-native speakers found a similarly increased preference for constructional meaning in more proficient speakers \citep{liang2002, sorting-italian}, suggesting commonalities in language acquisition between LMs and humans.

Our second case study is based on nonsense {``Jabberwocky''} sentences that nevertheless convey meaning when they are arranged in constructional templates \citep{johnson-goldberg}. We adapt the original priming experiment to LMs and show that RoBERTa is able to derive meaning from ASCs, even without any lexical cues. This finding offers counter-evidence to earlier claims that LMs are relatively insensitive to word order when constructing sentence meaning \citep{yu-ettinger, unnatural-inference}. Our source code and data are available at: \url{https://github.com/SPOClab-ca/neural-reality-constructions}.

\section{Psycholinguistic background}

\subsection{Construction grammar and ASCs}

Construction grammar is a family of linguistic theories proposing that all linguistic knowledge consists of {\em constructions}: pairings between form and meaning where some aspects of form or meaning are not predictable from their parts \citep{fillmore-et-all1988, kay-fillmore1999, goldberg1995, goldberg2006}. Common examples include idiomatic expressions such as {\em under the weather} (meaning { ``to feel unwell''}), but many linguistic patterns are constructions, including morphemes (e.g., {\em -ify}), words (e.g., {\em apple}), and abstract patterns like the ditransitive and passive. In contrast to lexicalist theories of argument structure, construction grammar rejects the dichotomy between syntax and lexicon. In contrast to transformational grammar, it rejects any distinction between surface and underlying structure.

We focus on a specific family of constructions for which there is an ample body of psycholinguistic evidence: argument structure constructions (ASCs). ASCs are constructions that specify the argument structure of a verb \citep{goldberg1995}. In the lexicalist, verb-centered view, argument structure is a lexical property of the verb, and the main verb of a sentence determines the form and meaning of the sentence \citep{chomsky1981, kaplan-bresnan1982, pollard-sag1987, levin-rappaport1995}. For example, {\em sneeze} is intransitive (allowing no direct object) and {\em hit} is transitive (requiring one direct object). However, lexicalist theories encounter difficulties with sentences like {\em ``he sneezed the napkin off the table''} since intransitive verbs are not permitted to have object arguments.

Rather than assuming multiple implausible senses for the verb {\em ``sneeze''} with different argument structures, \citet{goldberg1995} proposed that ASCs operate on an arbitrary verb, altering its argument structure while at the same time modifying its meaning. For example, the {\em caused-motion} ASC adds a direct object and a path argument to the verb {\em sneeze}, with the semantics of causing the object to move along the path. Other ASCs include the transitive, ditransitive, and resultative (Figure \ref{fig:cxn-examples}), which specify the argument structure of a verb and interact with its meaning in different ways.

\subsection{Psycholinguistic evidence for ASCs}

\begin{table*}[]
\small \centering
\begin{tabular}{l p{0.19\linewidth}p{0.19\linewidth}p{0.19\linewidth}p{0.19\linewidth}}
\hline
               & \textbf{Transitive}     & \textbf{Ditransitive}         & \textbf{Caused-motion}            & \textbf{Resultative}            \\ \hline
\textbf{Throw} & Anita threw the hammer. & Chris threw Linda the pencil. & Pat threw the keys onto the roof. & Lyn threw the box apart.        \\
\textbf{Get}   & Michelle got the book.  & Beth got Liz an invitation.   & Laura got the ball into the net.  & Dana got the mattress inflated. \\
\textbf{Slice} & Barbara sliced the bread. & Jennifer sliced Terry an apple. & Meg sliced the ham onto the plate. & Nancy sliced the tire open. \\
\textbf{Take}  & Audrey took the watch.  & Paula took Sue a message.     & Kim took the rose into the house. & Rachel took the wall down.      \\ \hline
\end{tabular}
\caption{Stimuli from \citet{bencini-goldberg}, consisting of a 4x4 design, with 4 different verbs and 4 different argument structure constructions.}
\label{table:bencini-goldberg-stimuli}
\end{table*}

{\bf Sentence sorting.} Several psycholinguistic studies have found evidence for argument structure constructions using experimental methods. Among these, \citet{bencini-goldberg} used a sentence sorting task to determine whether the verb or construction in a sentence was the main determinant of sentence meaning. 17 participants were given 16 index cards with sentences containing 4 verbs ({\em throw, get, slice}, and {\em take}) and 4 constructions ({\em transitive, ditransitive, caused-motion}, and {\em resultative}) and were instructed to sort them into 4 piles by overall sentence meaning (Table \ref{table:bencini-goldberg-stimuli}). The experimenters measured the deviation to a purely verb-based or construction-based sort, and found that on average, the piles were closer to a construction sort.

{\bf Non-native sentence sorting.} The same set of experimental stimuli was used with L2 (non-native) English speakers. \citet{gries-wulff} ran the experiment with 22 German native speakers, who preferred the construction-based sort over the verb-based sort, showing that constructional knowledge is not limited to native speakers. \citet{liang2002} ran the experiment on Chinese native speakers of 3 different English levels (46 beginner, 31 intermediate, and 33 advanced), and found that beginners preferred a verb-based sort, while advanced learners produced construction-based sorts similar to native speakers (Figure \ref{fig:human-lm-sorting-deviation}). Likewise, \citet{sorting-italian} found the same result in Italian native speakers with B1 and B2 English proficiency levels. Overall, these studies show evidence for ASCs in the mental representations of native and L2 English speakers alike, and furthermore, preference for constructional over verb sorting increases with increasing English proficiency.

{\bf Multilingual sentence sorting.} Similar sentence sorting experiments have been conducted in other languages, with varying results. \citet{sorting-german} ran a sentence sorting experiment in German with 40 participants and found that they mainly sorted by verb but rarely by construction. \citet{sorting-italian} ran an experiment with non-native learners of Italian (15 participants of B1 level and 10 participants of B2 level): both groups preferred the constructional sort, and similar to \citet{liang2002}, the B2 learners sorted more by construction than the B1 learners. \citet{sorting-spanish} ran an experiment in Spanish with 16 participants, and found approximately equal proportions of constructions and verb sort. In Italian and Spanish, some different constructions were substituted as not all of the English constructions had an equivalent in these languages; see the appendix for the complete set of stimuli in each language.

{\bf Priming.} Another line of psycholinguistic evidence comes from priming studies. Priming refers to the condition where exposure to a (prior) stimulus influences the response to a later stimulus \citep{pickering-priming}. \citet{bock90} found that participants were more likely to produce sentences of a given syntactic structure when primed with a sentence of the same structure; \citet{ziegler19} argued that \citet{bock90} did not adequately control for lexical overlap, and instead, they showed that the construction must be shared for the priming effect to occur, not just shared abstract syntax.

{\bf Novel verbs.} Even with unfamiliar words, there is evidence that constructions are associated with meaning. \citet{kaschak-glenberg} constructed sentences with novel denominal verbs and found that participants were more likely to interpret a transfer event when the denominal verb was used in a ditransitive sentence ({\em Tom crutched Lyn an apple}) than a transitive one ({\em Tom crutched an apple}).

\citet{johnson-goldberg} used a ``Jabberwocky'' priming task to show that abstract constructional templates are associated with meaning. Participants were primed with a nonsense sentence of a given construction (e.g., {\em He daxed her the norp} for the ditransitive construction), followed by a lexical decision task of quickly deciding if a string of characters was a real English word or a non-word. The word in the decision task was semantically congruent with the construction ({\em gave}) or incongruent ({\em made}); furthermore, they experimented with target words that were high-frequency ({\em gave}), low-frequency ({\em handed}), or semantically related but not associated with the construction ({\em transferred}). They found priming effects (faster lexical decision times) in all three conditions, with the strongest effect for the high-frequency condition, followed by the low-frequency and the semantically nonassociate conditions.

We adapt several of these psycholinguistic studies to LMs: the sentence sorting experiments in Case study 1, and the Jabberwocky priming experiment in Case study 2. We choose these studies because their designs allow for thousands of stimuli sentences to be generated automatically using templates, avoiding issues caused by small sample sizes from manually constructed sentences.

\section{Related work in NLP}

\subsection{Linguistic  probing of LMs}

Many studies have probed for various aspects of syntax in LSTMs and Transformer-based LMs. \citet{linzen2016} tested LSTMs on their ability to capture subject-verb agreement, using templates to generate test data. This idea was extended by BLiMP \citep{blimp}, a suite encompassing 67 linguistic phenomena, including filler-gap effects, NPI licensing, and ellipsis; \citet{hu-syntax-assessment} released a similar test suite. Template generation is a convenient method to construct stimuli exhibiting specific linguistic properties, but alternative approaches include CoLA \citep{cola}, which compiled an acceptability benchmark of sentences drawn from linguistic publications, and \citet{gulordava-colorless}, who perturbed natural sentences to study LMs' knowledge of agreement on nonsense sentences. We refer to \citet{linzen-baroni} for a comprehensive review of the linguistic probing literature.

So far, relatively few papers approached LM probing from a construction grammar perspective. \citet{cxgbert} probed for BERT's knowledge of constructions via a sentence pair classification task of predicting whether two sentences share the same construction. Their probe was based on data from \citet{dunn-data}, who used an unsupervised algorithm to extract plausible constructions from corpora based on association strength. However, the linguistic validity of these automatically induced constructions is uncertain, and there is currently no human-labelled wide-coverage construction grammar dataset in any language suitable for probing. Other computational work focused on a few specific constructions, such as identifying caused-motion constructions in corpora \citep{hwang-cm} and annotating constructions related to causal language \citep{dunietz-causal}. \citet{lebani-lenci} is the most similar to our work: they probed distributional vector space models for ASCs based on the Jabberwocky priming experiment by \citet{johnson-goldberg}.

\subsection{Psycholinguistic treatment of LMs}

Some recent probing studies adapted methods and data from psycholinguistic research, treating LMs as psycholinguistic participants. Using a cloze completion task, \citet{ettinger-psycholinguistic} found that BERT was less sensitive than humans at commonsense inferences and detecting role reversals, and fails completely at understanding negation. \citet{michaelov-n400} compared LM surprisals with the N400 (a measure of human language processing difficulty) across a wide range of conditions; \citet{layerwise-anomaly} used psycholinguistic stimuli and found that LMs exhibit different layerwise surprisal patterns for morphosyntactic, semantic, and commonsense anomalies. \citet{wilcox21} compared LM and human sensitivities to syntactic violations using a maze task to collect human reaction times. \citet{prasad19, misra-priming} investigated whether LMs are sensitive to priming effects like humans. The advantage of psycholinguistic data is that they are carefully constructed by expert linguists to test theories of language processing in humans; however, their small sample size makes it challenging to make statistically meaningful conclusions when the (oft-sparse) experimental stimuli are used to probe a  language model.

\section{Case study 1: Sentence sorting}
\label{sec:sentence-sorting}

\begin{figure*}[t]
    \centering
    \includegraphics[width=\linewidth]{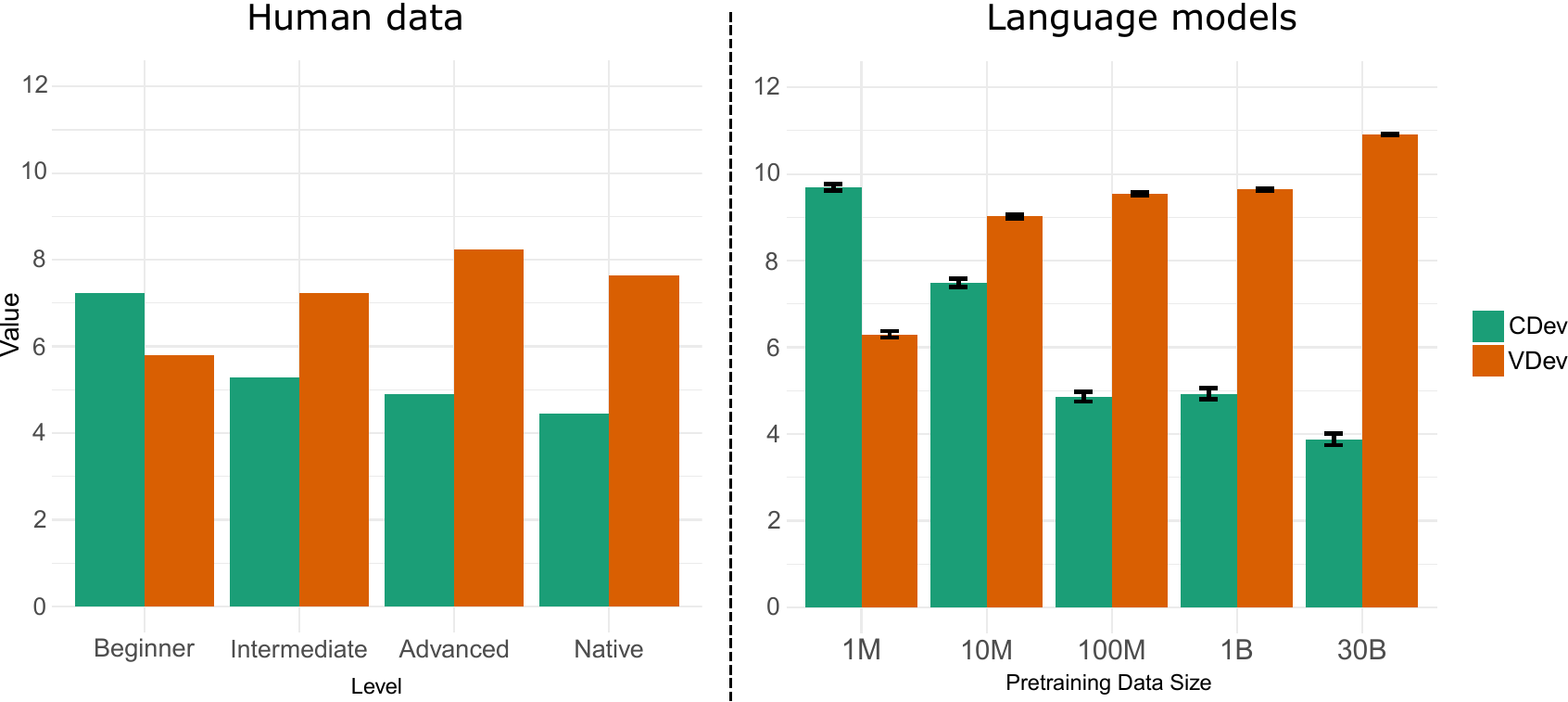}
    \caption{Sentence sorting results for humans and LMs, measured by deviation from pure construction and verb sort (CDev and VDev). Non-native human results are from \citet{liang2002}; native human results from \citet{bencini-goldberg}.\footnotemark LM results are obtained using MiniBERTas \citep{mini-bertas} and RoBERTa \citep{roberta} on templated stimuli. The MiniBERTa models use between 1M to 1B tokens for pretraining, while RoBERTa uses 30B tokens. Error bars indicate 95\% confidence intervals.}
    \label{fig:human-lm-sorting-deviation}
\end{figure*}
\footnotetext{\citet{bencini-goldberg} ran the sentence sorting experiment twice, so we take the average of the two runs.}

\begin{figure*}[t]
    \centering
    \includegraphics[width=\linewidth]{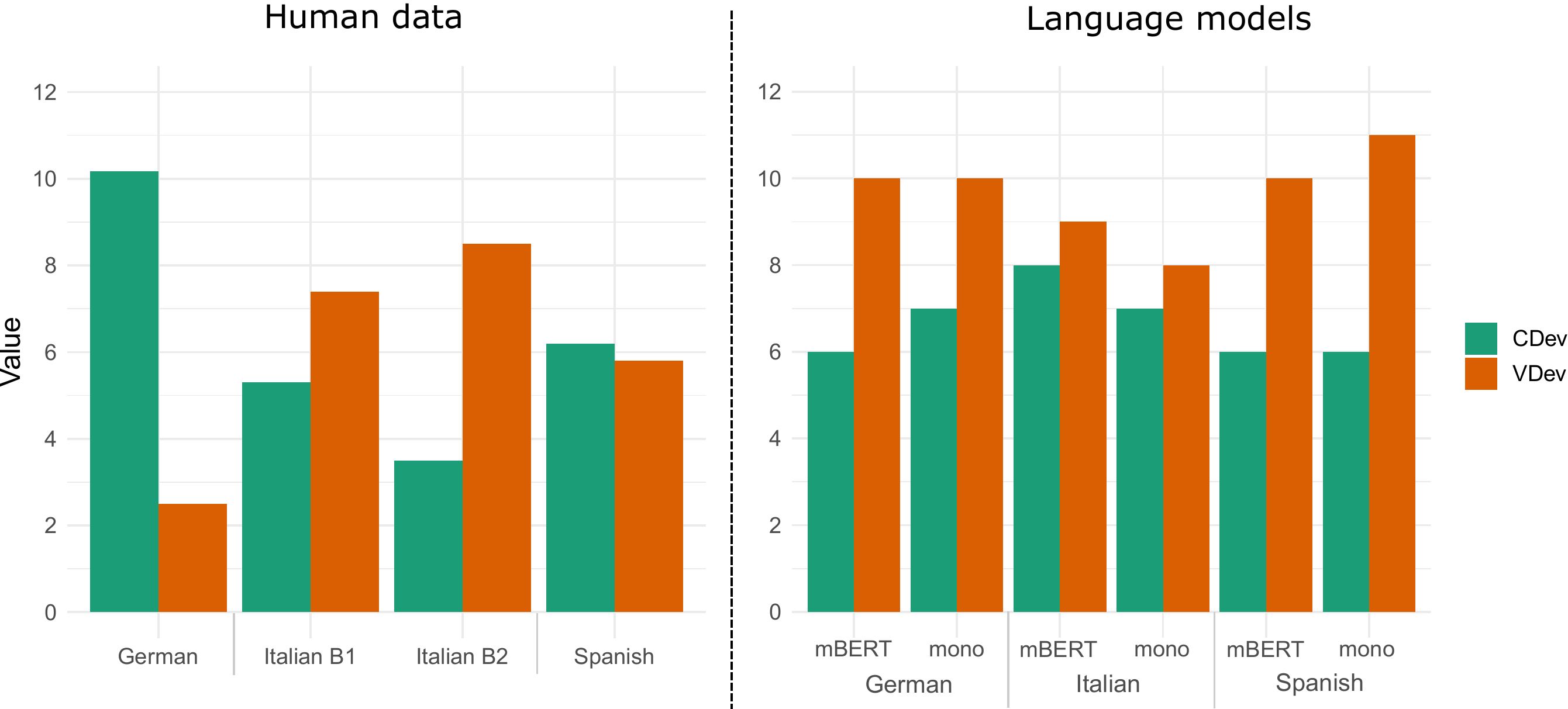}
    \caption{Multilingual sentence sorting results for German \citep{sorting-german}, Italian \citep{sorting-italian}, and Spanish \citep{sorting-spanish}. LM results are obtained using the same stimuli; we use both mBERT and a monolingual LM for each language.}
    \label{fig:multilang-sorting}
\end{figure*}

This section describes our adaptation of the sentence sorting experiments to Transformer LMs.

\subsection{Methodology}

{\bf Models.} To simulate varying non-native English proficiency levels, we use MiniBERTa models \citep{mini-bertas}, trained with 1M, 10M, 100M, and 1B tokens. We also use the base RoBERTa model \citep{roberta}, trained with 30B tokens. In other languages, there are no available pretrained checkpoints with varying amounts of pretraining data, so we use the mBERT model \citep{bert} and a monolingual Transformer LM in each language.\footnote{We use monolingual German and Italian models from \url{https://github.com/dbmdz/berts}, and the monolingual Spanish model from \citet{mono-spanish-lm}.} We obtain sentence embeddings for our models by taking the average of their contextual token embeddings at the second-to-last layer (i.e., layer 11 for base RoBERTa). We use the second-to-last because the last layer is more specialized for the LM pretraining objective and less suitable for sentence embeddings \citep{liu-transferability}.

{\bf Template generation.} We use templates to generate stimuli similar to the 4x4 design in the \citet{bencini-goldberg} experiment. To ensure an adequate sample size, we run multiple empirical trials. In each trial, we sample 4 random distinct verbs from a pool of 10 verbs that are compatible with all 4 constructions ({\em cut, hit, get, kick, pull, punch, push, slice, tear, throw}). We then randomly fill in the slots for proper names, objects, and complements for each sentence according to its verb, such that the sentence is semantically coherent, and there is no lexical overlap among the sentences of any construction. Table \ref{table:templated-bg-stimuli} in the appendix shows a set of template-generated sentences. In English, we generate 1000 sets of stimuli using this procedure; for other languages, we use the original stimuli from their respective publications.

{\bf Evaluation.} Similar to the human experiments, we group the sentence embeddings into 4 clusters (not necessarily of the same size) using agglomerative clustering by Euclidean distance \citep{sklearn}. We then compute the deviation to a pure construction and pure verb sort using the Hungarian algorithm for optimal bipartite matching. This measures the minimal number of cluster assignment changes necessary to reach a pure construction or verb sort, ranging from 0 to 12. Thus, lower construction deviation indicates that constructional information is more salient in the LM's embeddings.

\subsection{Results and interpretation}
 
Figure \ref{fig:human-lm-sorting-deviation} shows the LM sentence sorting results for English. All differences are statistically significant ($p < .001$). The smallest 1M MiniBERTa model is the only LM to prefer verb over construction sorting, and as the amount of pretraining data grows, the LMs increasingly prefer sorting by construction instead of by verb. This closely mirrors the trend observed in the human experiments.

The results for multilingual sorting are shown in Figure \ref{fig:multilang-sorting}. Both mBERT and the monolingual LMs consistently prefer constructional sorting over verb sorting in all three languages, whereas the results from the human experiments are less consistent.

Our results show that RoBERTa can generalize meaning from abstract constructions without lexical overlap. Only larger LMs and English speakers of more advanced proficiency are able to make this generalization, while smaller LMs and less proficient speakers derive meaning more from surface features like lexical content. This finding agrees with \citet{mini-bertas}, who found that larger LMs have an inductive bias towards linguistic generalizations, while smaller LMs have an inductive bias towards surface generalizations; this may explain the success of large LMs on downstream tasks. A small quantity of data (10M tokens) is sufficient for LMs to prefer the constructional sort, indicating that ASCs are relatively easy to learn: roughly on par with other types of linguistic knowledge, and requiring less data than commonsense knowledge \citep{zhang-billions, probing-across-time}.

We note some limitations in these results, and reasons to avoid drawing unreasonably strong conclusions from them. Human sentence sorting experiments can be influenced by minor differences in the experimental setup: \citet{bencini-goldberg} obtained significantly different results in two runs that only differed on the precise wording of instructions. In the German experiment \citep{sorting-german}, the author hypothesized that the participants were influenced by a different experiment that they had completed before the sentence sorting one. Given this experimental variation, we cannot attribute differences across languages to differences in their linguistic typology. Although LMs do not suffer from the same experimental variation, we cannot conclude statistical significance from the multilingual experiments, where only one set of stimuli is available in each language.

\section{Case study 2: Jabberwocky constructions}
\label{sec:jabberwocky-priming}

We next adapt the ``Jabberwocky'' priming experiment from \citet{johnson-goldberg} to LMs, and make several changes to the original setup to better assess the capabilities of LMs. Priming is a standard experimental paradigm in psycholinguistic research, but it is not directly applicable to LMs: existing methods simulate priming either by applying additional fine-tuning \citep{prasad19}, or by concatenating sentences that typically do not co-occur in natural text \citep{misra-priming}. Therefore, we instead propose a method to probe LMs for the same linguistic information using only distance measurements on their contextual embeddings.

\begin{figure}
    \centering
    \includegraphics[width=0.7\linewidth]{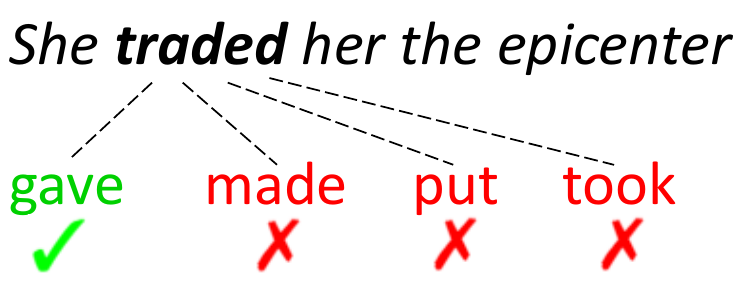}
    \caption{In our adapted Jabberwocky experiment, we measure the Euclidean distance from the Jabberwocky verb ({\em traded}) to the 4 prototype verbs, of which 1 is congruent ({\color{green}\cmark}) with the construction of the sentence, and 3 are incongruent ({\color{red}\xmark}).}
    \label{fig:jabberwocky-schema}
\end{figure}

\begin{table}[t]
\centering
\begin{tabular}{ll}
\hline
\textbf{Construction}          & \textbf{Template / Examples}             \\ \hline
\multirow{3}{*}{Ditransitive}  & S/he V-ed him/her the N.                  \\
                               & \textit{She traded her the epicenter.}    \\
                               & \textit{He flew her the donut.}           \\ \hline
\multirow{3}{*}{Resultative}   & S/he V-ed it Adj.                         \\
                               & \textit{He cut it seasonal.}              \\
                               & \textit{She surged it civil.}             \\ \hline
\multirow{3}{*}{Caused-motion} & S/he V-ed it on the N.                    \\
                               & \textit{He registered it on the diamond.} \\
                               & \textit{She awarded it on the corn.}    \\ \hline
\multirow{3}{*}{Removal}       & S/he V-ed it from him/her.                \\
                               & \textit{He declined it from her.}         \\
                               & \textit{She drove it from him.}           \\ \hline
\end{tabular}
\caption{Templates and example sentences for the Jabberwocky construction experiments. The templates are identical to the ones used in \citet{johnson-goldberg}, except that we use random real words instead of nonce words.}
\label{tab:jg-examples}
\end{table}

\subsection{Methodology}

{\bf Template generation.} We generate sentences for the four constructions randomly using the templates in Table \ref{tab:jg-examples}. Instead of filling nonce words like {\em norp} into the templates as in the original study, we take an approach similar to \citet{gulordava-colorless} and generate 5000 sentences for each construction by randomly filling real words of the appropriate part-of-speech into construction templates (Table \ref{tab:jg-examples}). This gives nonsense sentences like {\em ``She traded her the epicenter''}; we refer to these random words as {\em Jabberwocky words}. By using real words, we avoid any potential instability from feeding tokens into the model that it has never seen during pretraining. We obtain a set of singular nouns, past tense verbs, and adjectives from the Penn Treebank \citep{ptb}, excluding words with fewer than 10 occurrences.

{\bf Verb embeddings.} Our probing strategy is based on the assumption that the contextual embedding for a verb captures its meaning in context. Therefore, if LMs associate ASCs with meaning, we should expect the contextual embedding for the Jabberwocky verb to contain the meaning of the construction. Specifically, we measure the Euclidean distance to a {\em prototype} verb for each construction (Figure \ref{fig:jabberwocky-schema}). These are verbs that \citet{johnson-goldberg} selected whose meaning closely resembles the construction's meaning: {\em gave}, {\em made}, {\em put}, and {\em took} for the ditransitive, resultative, caused-motion, and removal constructions, respectively.\footnote{The reader may notice that the four constructions here are slightly different from \citet{bencini-goldberg}: the transitive construction is replaced with the removal construction in \citet{johnson-goldberg}.} We also run the same setup using lower frequency prototype verbs from the same study: {\em handed}, {\em turned}, {\em placed}, and {\em removed}.\footnote{\citet{johnson-goldberg} also included a third experimental condition using four verbs that are semantically related but not associated with the construction, but one of the verbs is very low-frequency ({\em ousted}), so we exclude this condition in our experiment.} As a control, we measure the Euclidean distance to the prototype verbs of the other three unrelated constructions.

The prototype verb embeddings are generated by taking the average across their contextual embeddings across a 4M-word subset of the British National Corpus (BNC; \citet{bnc}). We use the second-to-last layer of RoBERTa-base, and in cases where a verb is split into multiple subwords, we take the embedding of the first subword token as the verb embedding.

\subsection{Results and interpretation}

\begin{figure}
    \centering
    \includegraphics[width=\linewidth]{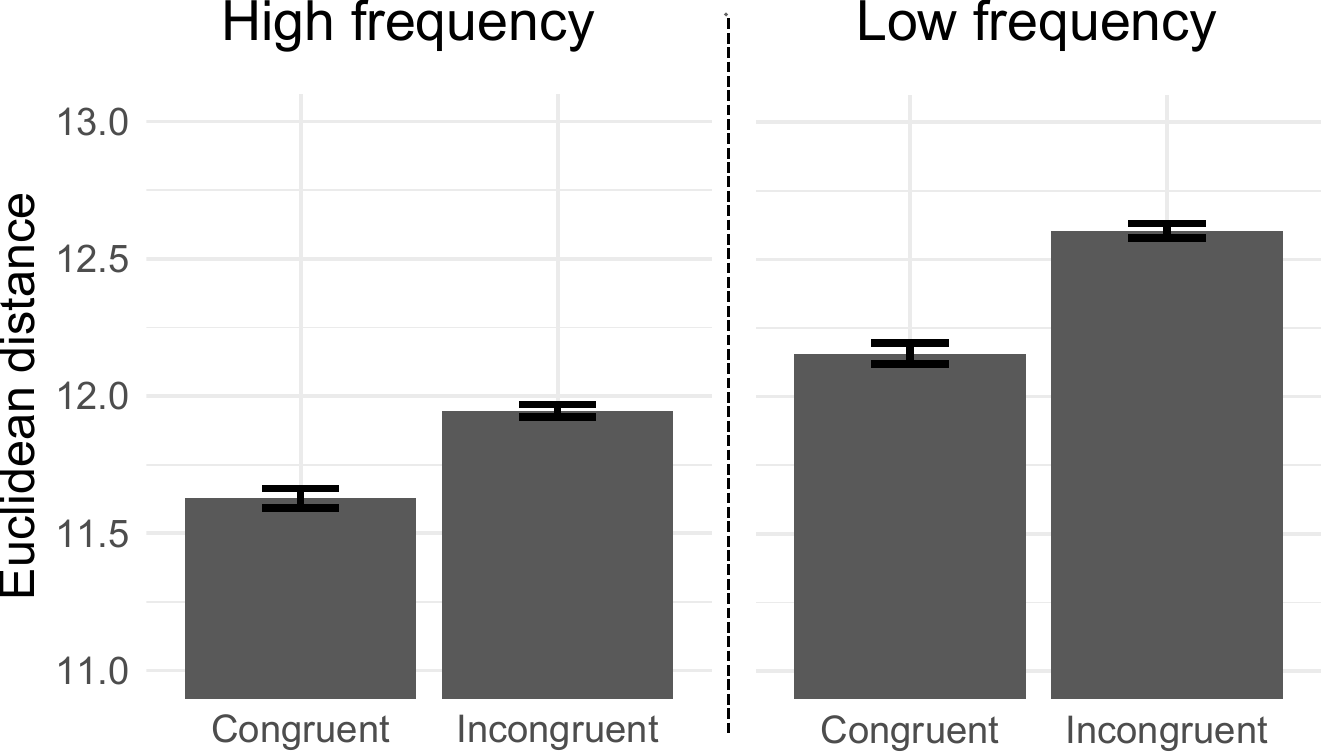}
    \caption{Euclidean distance between Jabberwocky and prototype verbs for congruent and incongruent conditions. Error bars indicate 95\% confidence intervals.}
    \label{fig:jg-barplot}
\end{figure}

\begin{figure*}
    \centering
    \includegraphics[width=\linewidth]{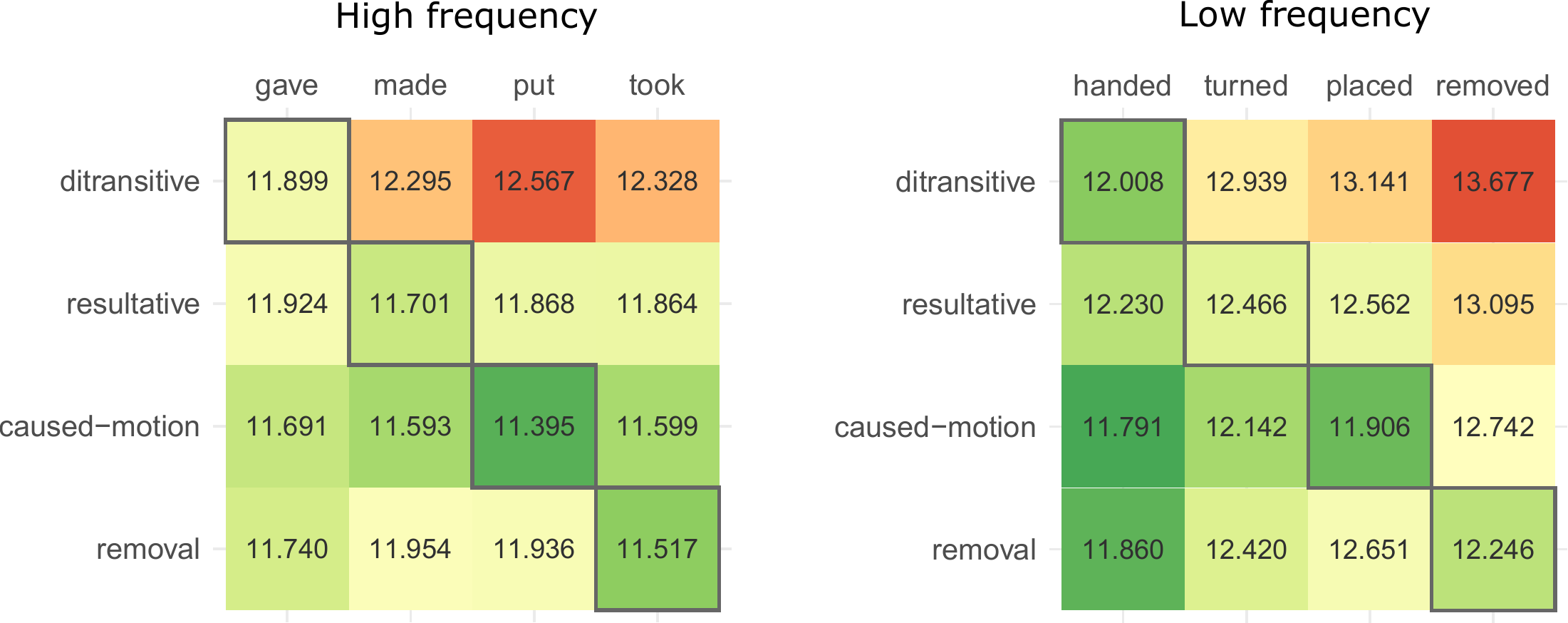}
    \caption{Mean Euclidean distance between Jabberwocky and prototype verbs in each verb-construction pair. Diagonal entries (gray border) are the congruent conditions; off-diagonal entries are incongruent.}
    \label{fig:jg-grid-4x4}
\end{figure*}

We find that the Euclidean distance between the prototype and Jabberwocky verb embeddings is significantly lower ($p < .001$) when the verb is congruent with the construction than when they are incongruent, and this is observed for both high and low-frequency prototype verbs (Figure \ref{fig:jg-barplot}). Examining the individual constructions and verbs (Figure \ref{fig:jg-grid-4x4}), we note that in the high-frequency scenario, the lowest distance prototype verb is always the congruent one, for all four constructions. In the low-frequency scenario, the result is less consistent: the congruent verb is not always the lowest distance one, although it is always still at most the second-lowest distance out of the four.

The main result holds for both high and low-frequency scenarios, but the correct prototype verb is associated more consistently in the high-frequency case. This agrees with \citet{wei-frequency}, who found that LMs have greater difficulty learning the linguistic properties of less frequent words. We also note that the Euclidean distances are higher overall in the low-frequency scenario, which is consistent with previous work that found lower frequency words to occupy a peripheral region of the embedding space \citep{layerwise-anomaly}.

\subsection{Potential confounds}

In any experiment, one must be careful to ensure that the observed patterns are due to the phenomenon under investigation rather than confounding factors. We discuss potential confounds arising from lexical overlap, anisotropy of contextual embeddings, and neighboring words.

{\bf Lexical overlap}. The randomized experiment design ensures that the Jabberwocky words cannot be lexically biased towards any construction, since each verb is equally likely to occur in every construction. Technically, the lexical content in the four constructions are not identical: i.e., words like {\em ``from''} (occurring only in the removal construction) or {\em ``on''} (in the caused-motion construction) may provide hints to the sentence meaning. However, the ditransitive and resultative constructions do not contain any such informative words, yet RoBERTa still associates the correct prototype verb for these constructions, so we consider it unlikely to be relying solely on lexical overlap. There is substantial evidence that RoBERTa is able to associate abstract constructional templates with their meaning without lexical cues. This result is perhaps surprising, given that previous work found that LMs are relatively insensitive to word order in compositional phrases \citep{yu-ettinger} and downstream inference tasks \citep{unnatural-inference, pham-out-of-order}, where their performance can be largely attributed to lexical overlap.

{\bf Anisotropy}. Recent probing work have found that contextual embeddings suffer from anisotropy, where embeddings lie in a narrow cone and have much higher cosine similarity than expected if they were directionally uniform \citep{how-contextual}. Furthermore, a small number of dimensions dominate geometric measures such as Euclidean and cosine distance, resulting in a degradation of representation quality \citep{kovaleva21, timkey21}. Since our experiments rely heavily on Euclidean distance, anisotropy is a significant concern. Following \citet{timkey21}, we perform standardization by subtracting the mean vector and dividing each dimension by its standard deviation, where the mean and standard deviation for each dimension is computed from a sample of the BNC. We observe little difference after standardization: in both the high and low frequency scenarios, the Euclidean distances are lower for the congruent than the incongruent conditions, by a similar margin compared to the original experiment without standardization. We also run standardization on the first case study, and find that the results remain essentially unchanged: smaller LMs still prefer verb sorting while larger LMs prefer construction sorting. Thus, neither of our experiments appear to be affected by anisotropy.

{\bf Neighboring words.} A final confounding factor is our assumption that RoBERTa's contextual embeddings represent word meaning, when in reality, they contain a mixture of syntactic and semantic information. Contextual embeddings are known to contain syntax trees \citep{hewitt-syntax} and linguistic information about neighboring words in a sentence \citep{klafka-ettinger}; although previous work did not consider ASCs, it is plausible that our verb embeddings leak information about the sentence's construction in a similar manner. If this were the case, the prototype verb embedding for {\em gave} would contain not only the semantics of transfer that we intended, but also information about its usual syntactic form\footnote{\citet{bresnan-gradience} estimated that 87\% of usages of the word {\em ``give''} occur in the ditransitive construction.} of {\em ``S gave NP1 NP2''}, and both would be captured by our Euclidean distance measurement. Controlling for this syntactic confound is difficult -- one could alternatively probe for transfer semantics without syntactic confounds using a natural language inference setup (e.g., whether the sentence entails the statement {\em ``NP1 received NP2''}), but we leave further exploration of this idea to future work.

\section{Conclusion}

We find evidence for argument structure constructions in Transformer language models from two separate angles: sentence sorting and Jabberwocky construction experiments. Our work extends the existing body of literature on LM probing by taking a constructionist instead of generative approach to linguistic probing. Our sentence sorting experiments identified a striking resemblance between humans' and LMs' internal language representations as LMs are exposed to increasing quantities of data, despite the differences between neural language models and the human brain. Our two studies suggest that LMs are able to derive meaning from abstract constructional templates with minimal lexical overlap. Both sets of experiments were inspired by psycholinguistic studies, which we adapted to fit the capabilities of LMs -- this illustrates the potential for future work on grounding LM probing methodologies in psycholinguistic research.

\section*{Acknowledgements}

YX is
funded through an NSERC Discovery Grant, a
SSHRC Insight Grant, and an Ontario ERA award.

\bibliography{naacl}
\bibliographystyle{acl_natbib}

\clearpage
\appendix

\section{Visualization of sentence sorting}

\begin{figure}
    \centering
    \includegraphics[width=0.8\linewidth]{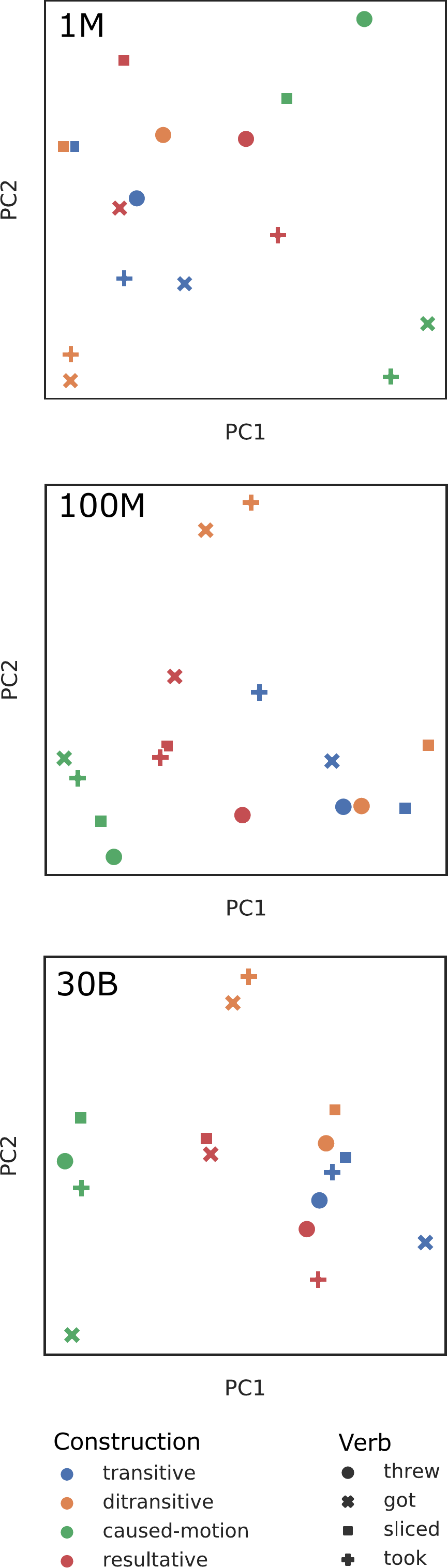}
    \caption{PCA plots of \citet{bencini-goldberg} sentence sorting using the 1M and 100M MiniBERTa models and RoBERTa-base (30B). Figure best viewed in color.}
    \label{fig:pca-plots}
\end{figure}

We use principal components analysis (PCA) to visualize the sentence sorting experiment for the MiniBERTa models (trained with 1M and 100M tokens) and RoBERTa-base (trained with 30B tokens). In RoBERTa, there is strong evidence of clustering based on constructions; the effect is unclear in the 100M model and nonexistent in the 1M model (Figure \ref{fig:pca-plots}). This visually confirms our quantitative evaluation based on the construction and verb deviation metrics (Figure \ref{fig:human-lm-sorting-deviation}).

\section{Additional experimental stimuli}

Table \ref{table:templated-bg-stimuli} shows an example set of template-generated stimuli for sentence sorting: we generate 1000 similar sets of 16 sentences to increase the sample size. We also present the sentence sorting stimuli for German (Table \ref{table:german-sorting-stimuli}), Italian (Table \ref{table:italian-sorting-stimuli}), and Spanish (Table \ref{table:spanish-sorting-stimuli}). German uses the same four constructions as English. Italian does not have the ditransitive construction but instead uses the prepositional dative construction to express transfer semantics. Spanish has no equivalents for the caused-motion and resultative constructions, so the authors in that experiment instead used the unplanned reflexive (expressing accidental or unplanned events), and the middle construction (expressing states pertaining to the subject).

\begin{table*}[]
\small \centering
\begin{tabular}{l p{0.19\linewidth}p{0.19\linewidth}p{0.19\linewidth}p{0.19\linewidth}}
\hline
               & \textbf{Transitive}     & \textbf{Ditransitive}         & \textbf{Caused-motion}            & \textbf{Resultative}            \\ \hline
\textbf{Slice}  & Harry sliced the bread. & Henry sliced Eric the box. & Sam sliced the ball onto the bed. & John sliced the book apart. \\
\textbf{Kick} & Thomas kicked the box. & Mike kicked Frank the ball. & Michael kicked the wall into the house. & James kicked the door open. \\
\textbf{Cut} & George cut the ball. & Adam cut Paul the tree. & Bill cut the box into the water. & Bob cut the bread apart. \\
\textbf{Get}  & Tom got the book. & Andrew got Steve the door. & Jack got the fridge onto the elevator. & David got the ball stuck. \\ \hline
\end{tabular}
\caption{Example of our 4x4 sentence sorting stimuli, similar to those by \citet{bencini-goldberg} in Table \ref{table:bencini-goldberg-stimuli}, but generated automatically using templates.}
\label{table:templated-bg-stimuli}
\end{table*}

\begin{table*}[]
\small \centering
\begin{tabular}{l p{0.19\linewidth}p{0.19\linewidth}p{0.19\linewidth}p{0.19\linewidth}}
\hline
               & \textbf{Transitive}     & \textbf{Ditransitive}         & \textbf{Caused-motion}            & \textbf{Resultative}            \\ \hline
\textbf{Werfen}  & Anita warf den Hammer.     & Berta warf Linda den Bleistift.      & Erika warf den Schlüsselbund auf das Dach. & Laura warf die Kisten auseinander.  \\
\textbf{Bringen} & Michelle brachte das Buch. & Simone brachte Lydia eine Einladung. & Emma brachte den Ball ins Netz.            & Leonie brachte die Stühle zusammen. \\
\textbf{Schneiden} & Karolin schnitt das Brot. & Luisa schnitt Paula einen Apfel. & Jennifer schnitt die Wurst auf den Teller. & Doris schnitt den Reifen auf. \\
\textbf{Nehmen}  & Maria nahm die Uhr.        & Sophia nahm Jasmin das Geld.         & Helena nahm die Rosen in das Haus.         & Theresa nahm das Plakat herunter.   \\ \hline
\end{tabular}
\caption{German sentence sorting stimuli, obtained from \citet{sorting-german}.}
\label{table:german-sorting-stimuli}
\end{table*}

\begin{table*}[]
\small \centering
\begin{tabular}{l p{0.19\linewidth}p{0.19\linewidth}p{0.19\linewidth}p{0.19\linewidth}}
\hline
               & \textbf{Transitive}     & \textbf{Prepositional Dative}         & \textbf{Caused-motion}            & \textbf{Resultative}            \\ \hline
\textbf{Dare} & Lauda dà un esame.	& Carlo dà una mela a Maria. & Luca dà una spinta a Franco.	& Paolo dà una verniciata di verde alla porta. \\
\textbf{Fare} & Mario fa una torta.	& Luigi fa un piacere a Giovanna.	& Fabio fa entrare la macchina in garage.	& Stefano fa bruciare il sugo. \\
\textbf{Mettere} & Annalisa mette la giacca.	& Riccardo mette il cappello al bambino.	& Silvia mette la penna nel cassetto.	& Filippo mette la casa in ordine. \\
\textbf{Portare}  & Linda porta lo zaino.	& Laura porta la pizza a Francesco.	& Michele porta il libro in biblioteca.	& Irene porta l'esercizio a termine. \\ \hline
\end{tabular}
\caption{Italian sentence sorting stimuli, obtained from \citet{sorting-italian}.}
\label{table:italian-sorting-stimuli}
\end{table*}

\begin{table*}[]
\small \centering
\begin{tabular}{l p{0.19\linewidth}p{0.19\linewidth}p{0.19\linewidth}p{0.19\linewidth}}
\hline
               & \textbf{Transitive}     & \textbf{Ditransitive}         & \textbf{Unplanned Reflexive}            & \textbf{Middle}            \\ \hline
\textbf{Romper}  & Carlos rompió el cristal. &	Alfonso le rompió las gafas a Pepe.&	A Juan se le rompieron los pantalones.&	La porcelana se rompe con facilidad. \\
\textbf{Doblar} & Felipe dobló el periódico.&	Pablo le dobló el brazo a Lucas.&	A Pedro se le dobló el tobillo.&	El aluminio se dobla bien. \\
\textbf{Acabar} & Leonardo acabó su tesis.&	Tomás le acabó la pasta de dientes a Santi.&	A Luis se le acabaron los cigarrillos.&	Las carreras de 10 km se acaban sin problemas. \\
\textbf{Cortar}  & Isidro cortó el pan.&	Jorge le cortó el paso a Yago.&	A Ignacio se le cortó la conexión.&	Esta tela se corta muy bien.  \\ \hline
\end{tabular}
\caption{Spanish sentence sorting stimuli, obtained from \citet{sorting-spanish}.}
\label{table:spanish-sorting-stimuli}
\end{table*}

\end{document}